\documentclass[sigconf]{acmart}
\usepackage{indentfirst}
\usepackage{bm}
\usepackage{booktabs}
\usepackage{multirow}
\usepackage{balance}
\usepackage[ruled,linesnumbered]{algorithm2e}
\usepackage{amsmath}
\usepackage{xcolor}
\usepackage{enumitem}

\AtBeginDocument{%
  \providecommand\BibTeX{{%
    \normalfont B\kern-0.5em{\scshape i\kern-0.25em b}\kern-0.8em\TeX}}}

\setcopyright{acmcopyright}
\copyrightyear{2021}
\acmYear{2021}
\setcopyright{acmcopyright}
\acmConference[MM '21] {Proceedings of the 29th ACM International Conference on Multimedia}{October 20--24, 2021}{Virtual Event, China.}
\acmBooktitle{Proceedings of the 29th ACM International Conference on Multimedia (MM '21), Oct. 20--24, 2021, Virtual Event, China}
\acmPrice{15.00}
\acmISBN{978-1-4503-8651-7/21/10}
\acmDOI{10.1145/3474085.3475279}

\begin{document}
\fancyhead{}

\title{Adaptive Normalized Representation Learning for Generalizable Face Anti-Spoofing}




\author{
    Shubao Liu
}
\affiliation{
    East China Normal University
}
\email{
    shubaoL@stu.ecnu.edu.cn
}
\authornote{Equal Contribution}

\author{
    Ke-Yue Zhang
}
\affiliation{
    Youtu Lab, Tencent, Shanghai
}
\email{
    zkyezhang@tencent.com
}
\authornotemark[1]

\author{
    Taiping Yao
}
\affiliation{
    Youtu Lab, Tencent, Shanghai
}
\email{
    taipingyao@tencent.com
}
\authornotemark[1]

\author{
    Mingwei Bi
}
\affiliation{
    Youtu Lab, Tencent, Shanghai
}
\email{
    mingweibi@tencent.com
}

\author{
    Shouhong Ding
}
\affiliation{
    Youtu Lab, Tencent, Shanghai
}
\email{
    ericshding@tencent.com
}
\authornote{Corresponding Author}

\author{
    Jilin Li
}
\affiliation{
    Youtu Lab, Tencent, Shanghai
}
\email{
    jerolinli@tencent.com
}

\author{
    Feiyue Huang
}
\affiliation{
    Youtu Lab, Tencent, Shanghai
}
\email{
    garyhuang@tencent.com
}

\author{
    Lizhuang Ma
}
\affiliation{
    East China Normal University
    Shanghai Jiao Tong University
}
\email{
    lzma@cs.ecnu.edu.cn
}
\authornotemark[2]


\begin{abstract}
With various face presentation attacks arising under unseen scenarios, face anti-spoofing (FAS) based on domain generalization (DG) has drawn growing attention due to its robustness. 
Most existing methods utilize DG frameworks to align the features to seek a compact and generalized feature space.
However, little attention has been paid to the feature extraction process for the FAS task, especially the influence of normalization, which also has a great impact on the generalization of the learned representation.
To address this issue, we propose a novel perspective of face anti-spoofing that focuses on the normalization selection in the feature extraction process.
Concretely, an Adaptive Normalized Representation Learning (ANRL) framework is devised, which adaptively selects feature normalization methods according to the inputs, aiming to learn domain-agnostic and discriminative representation.
Moreover, to facilitate the representation learning, Dual Calibration Constraints are designed, including Inter-Domain Compatible loss and Inter-Class Separable loss, which provide a better optimization direction for generalizable representation.
Extensive experiments and visualizations are presented to demonstrate the effectiveness of our method against the SOTA competitors.

\end{abstract}




\maketitle

\section{Introduction}
\begin{figure}[t!]
	\centering
	\includegraphics[width=1.0\linewidth]{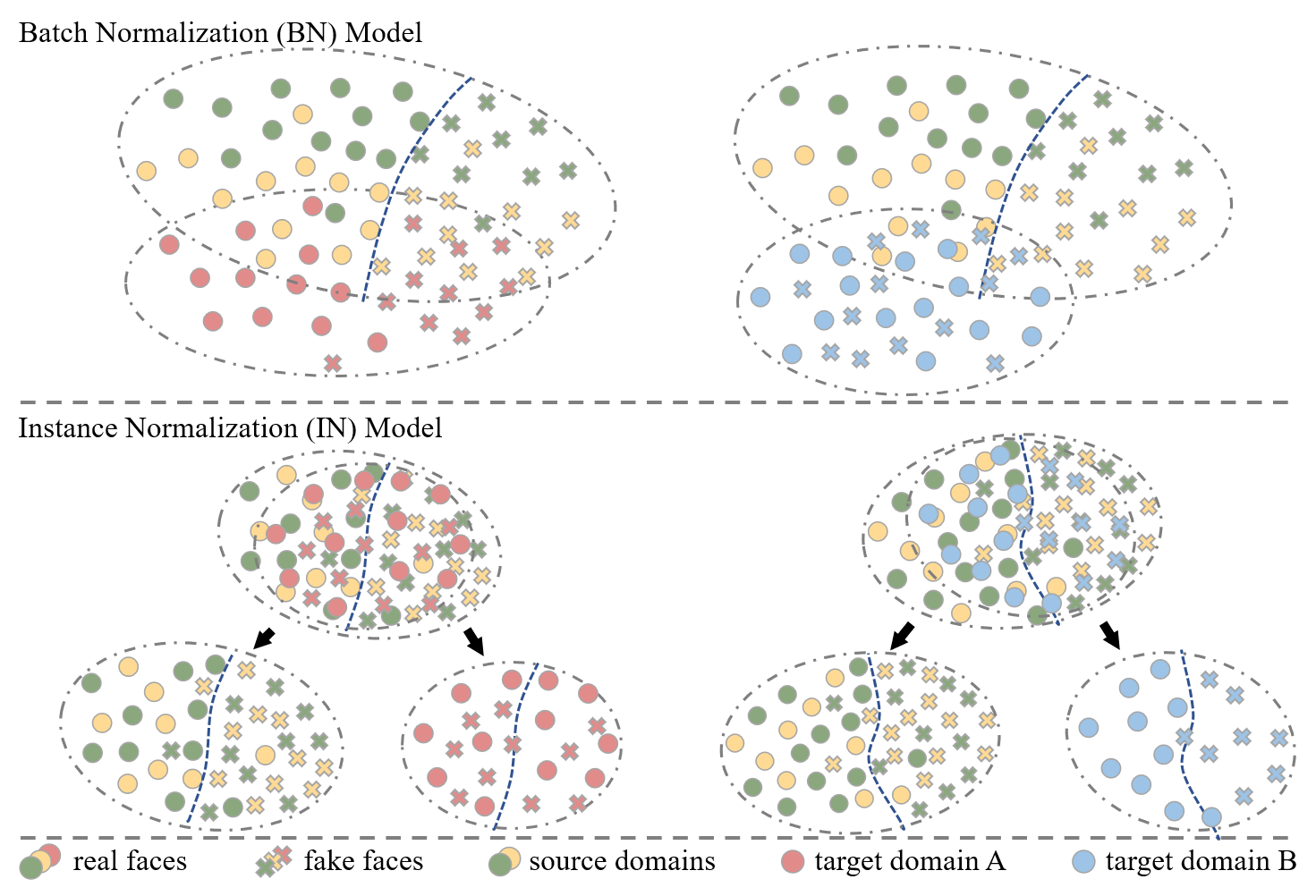}
	\caption{\textbf{The illustration of results on different target domains of models with BN or IN trained on the same source domains.} 
	It is obvious that both normalization methods have limitations.
    In the left part, for the red target domain A, BN outperforms IN. In the right part, for the blue target domain B, IN achieves better performance than the BN.
	}
	\label{illustration}
\end{figure}
Since various face presentation attacks arise, the safety of face recognition systems has become a critical public concern. To tackle this problem, researchers put forward many face anti-spoofing (FAS) methods, which initially leverage hand-craft features, such as LBP~\cite{LBP00,LBP01,LBP03} and HOG~\cite{HoG00,HoG01} to extract the spoof related texture for distinguishing the real and fake faces. Coming into the era of deep
learning, some methods~\cite{DeepBinary00,DeepBinary01,DeepBinary02,DeepBinary03,zhang2021structure} utilize CNN to detect face attacks via its strong representation abilities.
Although these methods have achieved extraordinary performance under intra-dataset testing scenarios, they all suffer from significant performance degradation when testing under cross-dataset scenarios. 
The reason behind this degradation is that these methods just fit on the training data with biased features and ignore the domain gap between source domains and unseen target domains, leading to poor generalization.

To tackle this problem, several methods introduce domain generalization into FAS tasks. Specifically, they~\cite{DG00,DG01,DG02} always map features from multiple source domains into a common feature space for generalizable representation, which can transfer well to unseen target domains.
While these methods just focus on the alignment of final features, but overlook the process of feature extraction.
Concretely, they always utilize the common module, \textit{i.e.,} CNN-BN-ReLU block, and we argue that the normalization is vital to this issue.
As shown in Figure~\ref{illustration}, we train the models with batch normalization (BN) or instance normalization (IN) on the same source domains and illustrate the results on different unseen target domains.
In the left part of Figure~\ref{illustration}, we test on target domain A, where the model with BN outperforms the one with IN. However, in the right part, IN achieves better performance than BN on target domain B. This can be attributed to the different properties of BN and IN. 
When the unseen target domain has little domain shift compared with source domains, BN maintains high performance. Since BN is easily affected by the domain information, its performance degrades encountering the large domain shift. While IN eliminates specific style information of each sample via using its own statistics, it is more tolerant of domain shift to improve the performance.

Since both normalizations have limitations, researchers have proposed some normalization combination methods to handle more situations, including the non-parametric methods~\cite{BIN00, Pan2018TwoAO} and parametric methods~\cite{BIN01,BIN02} . 
However, since samples in FAS tasks are different in scenes, lighting, \textit{etc.}, it is not suitable for them to utilize a shared parameter for combination without considering the uniqueness of each sample, which may lead to performance degradation.
In addition, they may still overfit to the source domains, resulting in the parameters not working well on unseen target domains.

To address the above limitations, we propose a novel framework, Adaptive Normalized Representation Learning (ANRL), to obtain a domain-agnostic and discriminative representation for FAS via adaptively selecting features from different normalization. 
Specifically, we put forward Adaptive Feature Normalization Module (AFNM) to estimate sample-wise factors for the fusion of IN and BN, which is aware of the distinction among samples. 
Furthermore, to assist AFNM in learning sample-wise factors, Dual Calibrated Constraints are proposed, containing Inter-Domain Compatible loss and Inter-Class Separable loss. 
The former loss aims to align the feature distribution of different domains, while the latter one is utilized to enlarge the margin between real samples and fake ones.
They cooperatively provide a better direction to update AFNM via meta-learning for generalization.

The main contributions of this work are summarized as follows:

$\bullet$ From a novel perspective, we propose to adaptively select different normalized features to obtain domain-agnostic and discriminative representation for generalizable face anti-spoofing.

$\bullet$ We propose Dual Calibrated Constraints, including Inter-Domain Compatible loss and Inter-Class Separable loss, to guide AFNM in estimating sample-wise factors for better generalization.

$\bullet$ Extensive experiments and visualizations are presented to reveal the role of adaptive normalization, which demonstrates the effectiveness of our method against state-of-the-art competitors.

\section{Related Work}
\subsection{Face Anti-Spoofing}
In recent years, researchers have made great progress in the face anti-spoofing area. 
The development is divided into two stages. 
Early researchers mainly utilized handcrafted feature descriptors, such as LBP~\cite{LBP00,LBP01,LBP03}, HOG~\cite{HoG00,HoG01}, SIFT~\cite{SIFT00} and then trained a traditional classifier for judgment. 
With the rise of deep learning, ~\cite{DeepBinary00,DeepBinary01,DeepBinary02,DeepBinary03,zhang2021structure,wang2020deep} regarded the face anti-spoofing as a binary classification task and leveraged CNN to solve it. However, such supervision may lead the model to a local optimum. 
To avoid easily overfitting, The methods in~\cite{additionInfo01,additionInfo00,additionInfo02,additionInfo04, Zhang2021AuroraGR} utilized additional supervisions, such as depth map~\cite{yu2021revisiting}, reflection map and r-ppg signal~\cite{lin2019face,yu2019remote1,niu2020video}, to boost the performance. Based on auxiliary information, the method~\cite{disentangle01,STCN} regularized features from the perspective of disentanglement.
Some methods~\cite{yu2021dual,yu2020multi} put forward specific convolution operators to extract spoof cues, such as CDCN~\cite{CDCN}, BCN~\cite{BCN}. 
The above methods got high performance under the intra-dataset setting, where the testing data comes from a similar distribution of training data. 
However, the performance under the cross-dataset setting always drops significantly.
To improve the performance of generalization, several methods~\cite{qin2019learning} introduced domain generalization (DG) into the face anti-spoofing area.
While MADDG~\cite{DG00} aligned all samples equally to learn a generalized feature space, SSDG~\cite{DG02} only aligned real samples from different datasets but not fake ones. Moreover, DRDG~\cite{liu2021dual} proposed to align samples with adaptative weights according to their domain information. Afterward, D2AM~\cite{Chen2021GeneralizableRL} was proposed to settle a more challenging generalizable scenario in the real world where domain labels are unknown.
Based on meta-learning, RFM~\cite{DG01} was updated via a more generalized optimization direction to get a robust classifier.
Although these methods gained a better generalization via DG frameworks, they overlooked the process of feature extraction, which were not sufficient for FAS domain generalization.

\begin{figure*}[t!]
	\centering
	\includegraphics[width=1.0\linewidth]{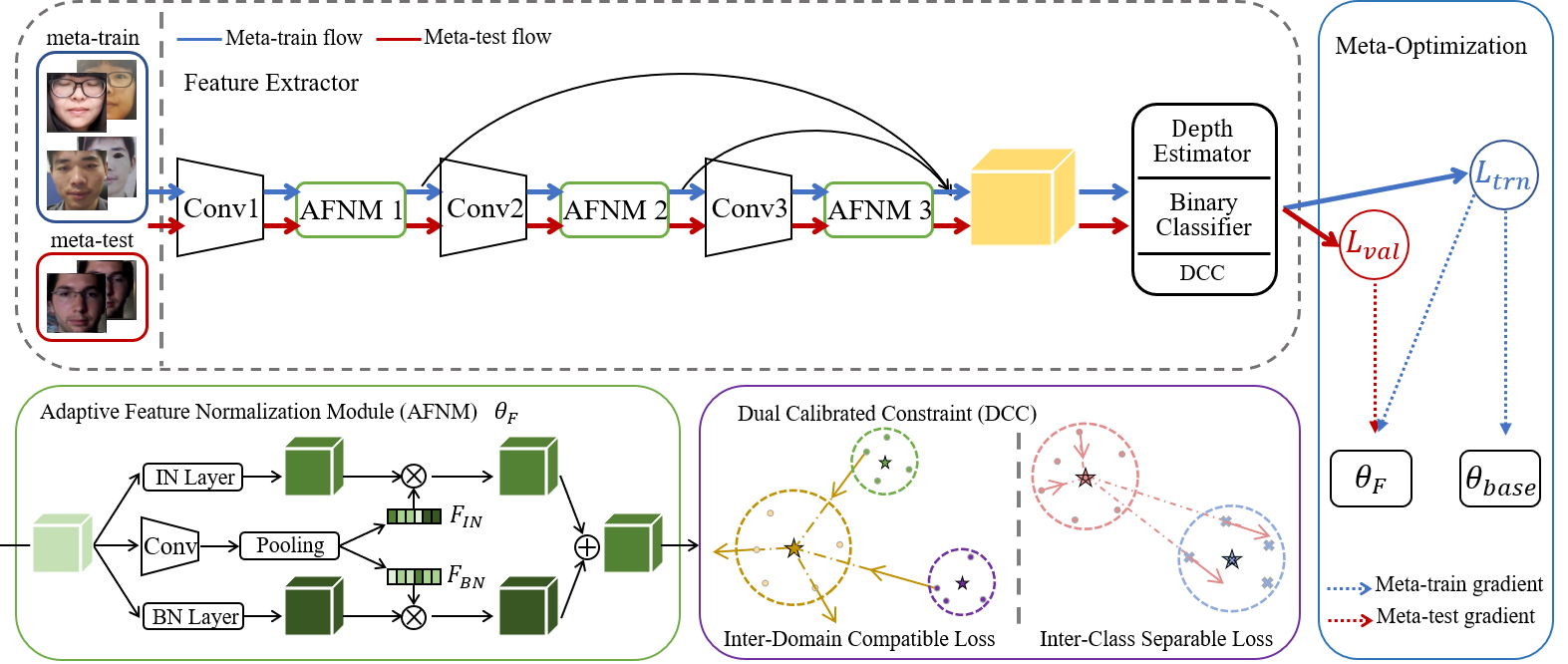}
	\caption{\textbf{The Adaptive Normalized Representation Learning (ANRL) framework commits to obtaining domain-agnostic and discriminative representations for FAS.} 
    Specifically, since the domain information contained in samples varies from each other, we utilize Adaptive Feature Normalization Module (AFNM) to adaptively fuse normalized features from BN and IN. 
    Moreover, Dual Calibrated Constraints (DCC) are introduced to aggregate the multiple source samples of the same class and split real samples from fake ones, providing a more generalizable direction to update AFNM via meta-learning.}
	\label{fig:framework}
\end{figure*}

\subsection{Batch-Instance Normalization}
Normalization techniques are significant parts of deep learning for better optimization and regularization. 
Several methods have attempted to combine batch normalization (BN) and instance normalization (IN) to improve generalization, which are divided into two categories: non-parametric methods and parametric methods.
The non-parametric methods~\cite{BIN00, Pan2018TwoAO} focused on strategies to replace BN with IN for some layers or add IN into specific locations. 
The parametric methods~\cite{BIN01,BIN02} learned a fixed parameter for the balance of BN and IN. 
For example, ~\cite{BIN01} introduced an effective batch-instance normalization layer through a simple training strategy, in which BN and IN were balanced with learnable parameters. 
Although they have improved performance, considering the distinction among samples, utilizing shared parameters may cause performance degradation on unseen target domains.

\section{Proposed Method}
\subsection{Overview}
Since the features extracted by BN are intrinsically fragile for domain shift and features extracted by IN may lose some discriminative information, both of them are limited to generalize on unseen target domains. Therefore, as illustrated in Figure \ref{fig:framework}, we propose Adaptive Normalized Representation Learning (ANRL) to acquire generalizable features, which not only retain discriminative information for spoof detection but also remove the domain variations. 
Specifically, since some samples are little domain-biased and some ones are large domain-biased, we devise Adaptive Feature Normalization Module (AFNM) to adaptively combine features from BN and IN with the most suitable sample-wise factors. 
To guide the estimation of generalizable factors, we propose Dual Calibrated Constraints (DCC), including Inter-Domain Compatible loss and Inter-Class Separable loss. 
Concretely, the former is dual-directionally designed to not only reduce inter-domain distance, but also scatter samples from the same domain to further promote the mixing of domains, and the latter is leveraged to separate real samples and fake ones with a larger margin.
Finally, combined with meta-learning, DCC provides a more generalizable direction to optimize AFNM, leading to constructing a compact and distinguishing normalized representation space.

\subsection{Adaptive Feature Normalization Module}
Though BN extracts discriminative information on source domains for anti-spoofing, it may not work when transferring to unseen target domains with excessive domain variations. 
To overcome the limitation, IN turns out to be an effective scheme for its capability to remove domain discrepancy. 
Inspired by batch-instance normalization (BIN)~\cite{BIN01} and each sample with different domain variations, we design Adaptive Feature Normalization Module (AFNM) to incorporate features extracted by BN and IN with adaptive balance factors customized for each sample.

Let $X \in \mathbb{R}^{C \times H \times W}$ be the feature map of the input image, where $C$, $H$ and $W$ denote channel, height and width respectively. 
We utilize BN and IN to acquire normalized representation respectively, which are denoted by $X^{BN}$ and $X^{IN}$. 
Since the balance factors should be suitable for each sample, we mine information from their corresponding features to generate balance factors. 
First, we leverage global average pooling (gp) to generate channel-wise statistics as $S\in\mathbb{R}^{C}$, which can be seen as the global information of feature map $X$. 
Concretely, the $c-$th channel of $S$ is calculated by shrinking $X$ through spatial dimensions $H\times W$ as follows:
\begin{equation}
    S_c = F_{gp}(X)=\frac{1}{H \times W}\sum_{i=1}^H\sum_{j=1}^W X_c(i,j).
\end{equation}
Then, a compact representation $Z\in\mathbb{R}^{d}$ is created to guide the adaptive selection, which is achieved by a fully connected (fc) layer to improve the efficiency:
\begin{equation}
    Z = F_{fc}(S)=\delta(WS),
\end{equation}
where $\delta$ is the ReLU function and $W\in\mathbb{R}^{d\times C}$ is the weight matrix of fc layer. Since $X^{BN}$ and $X^{IN}$ focus on different information, we respectively utilize soft attention across channels to adaptively select useful information as follows:
\begin{equation}
    B = \sigma(W_B Z), 
    \quad I = \sigma(W_I Z),
\end{equation}
where $\sigma$ is the Sigmoid function, $W_B, W_I \in \mathbb{R}^{C\times d}$ denote weight matrix for BN and IN respectively. 
Afterwards, the $c$-th element $\alpha_c$ of normalized balance factor $\alpha \in \mathbb{R}^{C}$ can be introduced:
\begin{equation}
    \alpha_c = \frac{B_c}{B_c+I_c}.
\end{equation}
The final feature map $Y$ is obtained through the attention weights on different feature maps from BN and IN:
\begin{equation}
    Y_c = \alpha_c X_c^{BN} + (1-\alpha_c) X_c^{IN},
\end{equation}
where $Y=[Y_1, Y_2, \dots, Y_c], Y_c \in \mathbb{R}^{H \times W}$.
Since the sample-wise $\alpha$ is generated according to the different domain information of each sample, AFNM can automatically select to rely more on IN or BN for generalizable representation.

\subsection{Dual Calibrated Constraints}

To guide AFNM in estimating better balance factors for fusing features from BN and IN, we propose Dual Calibrated Constraints (DCC). Different from the commonly used triplet losses for FAS tasks, which only construct triplets based on classes, DCC constrains features more comprehensively from the perspective of domain and class simultaneously.

\noindent \textbf{Inter-Domain Compatible Loss.} Since BN is vulnerable to domain differences, we disarrange inter-domain distributions to narrow gaps among multi-source domains. 
For the purpose, dual-directional Inter-Domain Compatible (IDC) loss is devised to pull samples of different domains close and scatter samples of the same domain away. Concretely, supposing there are $K$ source domains in one local mini-batch, we introduce the local centroid in domain $k$ as $C_k = \frac{1}{N_k}\sum_{i=1}^{N_k} O_i^k$, where $C_k$ denotes the centroid of domain $k$, $N_k$ is the number of samples and $O_i^k$ denotes the extracted feature of $i$-th sample in domain k. Moreover, to estimate more accurate centroid, we calculate the global centroid through different batches with the momentum update mechanism as $\widehat{{C_k}}=\gamma \widehat{{C_k}} + (1-\gamma) {C_k}$, where $\gamma$ is the momentum factor. 

Then, for the specific domain $k$, we calculate the intra-domain distance $D_{sd}^k$ between samples and the related global centroid:
\begin{equation}
    D_{sd}^k = \frac{1}{N_k}\sum_{i=1}^{N_k}(O_i^k-\widehat{C_k})^2,
\end{equation}
Moreover, we calculate the inter-domain distance $D_{dd}^k$ between samples in domain $k$ and global centroids of other domains:
\begin{equation}
    D_{dd}^k = \frac{1}{K-1}\sum_{m\neq k}^K\frac{1}{N_{k}}\sum_{i=1}^{N_{k}}(O_i^k-\widehat{C_{m}})^2,
\end{equation} 
Lastly, the IDC loss $\mathcal{L}_{IDC}$ is defined as:
\begin{equation}
    \mathcal{L}_{IDC} = \sum_{k=1}^K(D_{dd}^k - D_{sd}^k).
\end{equation}

The dual-directional designed IDC loss not only reduces inter-domain distance, but also scatters samples from intra-domain to further promote the mixing of different domains. By doing so, we mitigate the corruption of BN stemming from the domain bias for domain-agnostic representation.

\noindent \textbf{Inter-Class Separable Loss.} 
Due to IN may eliminate some information beneficial to the FAS task, it is necessary to keep discriminative information via enlarging the margin between real samples and fake ones.
Therefore, we design Inter-Class Separable (ICS) loss to gather samples of the same class together and keep samples of different classes apart. 
Following the calculation of $\widehat{C_k}$, we attain the centroid of real class $\widehat{C_r}=\frac{1}{N_r}\sum_{i=1}^{N_r} O_i^r$ and fake class $\widehat{C_f}=\frac{1}{N_f}\sum_{i=1}^{N_f} O_i^f$ respectively, where $N_r$ is the number of real samples and $N_f$ is the number of fake ones.

Because of the variety of attacks, the distribution of fake samples may be inherently scattered, and it will bring negative effects to forcibly aggregate them together~\cite{DG02}. 
However, the distribution of real ones is relatively stable, suitable for improving compactness. 
Therefore, we only compute intra-class distance $D_{rr}$ from real samples to $\widehat{C_r}$ as follows:
\begin{equation}
    D_{rr} = \frac{1}{N_r}\sum_{i=1}^{N_r}(O_i^r-\widehat{C_r})^2.
\end{equation}
Furthermore, the inter-calss distance $D_{rf}$ between real samples to $\widehat{C_f}$ is defined as:
\begin{equation}
    D_{rf} = \frac{1}{N_r}\sum_{i=1}^{N_r}(O_i^r-\widehat{C_f})^2,
\end{equation}
In the same way, we can calculate $D_{fr}$ as the distance between fake samples and $\widehat{C_r}$. 
Finally, ICS loss $\mathcal{L}_{ICS}$ is:
\begin{equation}
    \mathcal{L}_{ICS} = D_{rr} - D_{rf} - D_{fr}.
\end{equation}

Via ICS loss, the model is forced to not only mine more distinguishing features to separate real samples from fake ones, but also tighten differences within the real class, both of which contribute to the discriminative representation for face anti-spoofing.

\subsection{Training Strategy}
In this section, we introduce other loss modules and the optimization schedule of our framework in detail, as shown in Algorithm~\ref{optimization_strategy}.
Following the conventional FAS methods, we adopt Feature Extractor, Depth Estimator and Binary Classifier to settle the face anti-spoofing issue.
Depth Estimator estimates the facial depth maps for live faces and zero maps for spoof faces to facilitate the learning of Feature Extractor.
The pseudo-depth maps for live faces are calculated by PRNet~\cite{pseudo-depth}. 
We utilize $\mathcal{L}_{Dep}$ to update Depth Estimator.
\begin{equation}
    \mathcal{L}_{Dep} = \sum_{(x_i,dep_i)} \left\| Dep(Ext(x_i)) - dep_i \right\|_{2}^{2}
\end{equation}
Binary Classifier detects the spoof faces from the real ones, which is optimized via $\mathcal{L}_{Cls}$.
\begin{equation}
    \mathcal{L}_{Cls} = -\sum_{(x_i,y_i)} y_i log(BC(Ext(x_i)))
\end{equation}
And the output of the Binary Classifier is the only metric for evaluating the results in this paper.

Since meta-learning has shown its potential on promoting generalization through the simulation of real domain shifts among multi-source domains, we utilize its learning strategy to optimize AFNM, improving the generalization of factor estimation for IN and BN. 
It is noted that only AFNM is updated via meta-learning strategy, while the other parameters in the base model including Feature Extractor, Binary Classifier and Depth Estimator follow the normal training process. 
Formally, we denote $\theta_{F}$ as the parameters of AFNM and $\theta_{base}$ as the parameters of the base model. 

For clearly describing the whole updating process of $\theta_{F}$ and $\theta_{base}$, we elaborate on one complete iteration in detail.

\noindent \textbf{Normal Train.} 
we utilize $\mathcal{L}_{base}$ calculated with batches sampled from all domains $D$ to update $\theta_{base}$ for attack detection:
\begin{equation}
\begin{split}
    &\mathcal{L}_{base}(\theta_{base},\theta_{F})= \sum_{D} \mathcal{L}_{Cls} + \mathcal{L}_{Dep} \\
    &\theta_{base} \leftarrow \theta_{base} - \beta_1 \nabla_{\theta_{base}} \mathcal{L}_{base}(\theta_{base},\theta_{F})
\end{split}
\end{equation}

\noindent \textbf{Meta-Train.}
Following the conventional meta-learning settings, we firstly split source domains into meta-train domains $D_{trn}$ and meta-test domains $D_{val}$.
Then we input batches sampled from $D_{trn}$ to the networks for calculating $\mathcal{L}_{trn}$,

\begin{equation}
    \mathcal{L}_{trn}(\theta_{base},\theta_{F})= \sum_{D_{trn}} \mathcal{L}_{Cls} + \mathcal{L}_{Dep}+ \lambda_1 \mathcal{L}_{IDC} + \lambda_2 \mathcal{L}_{ICS}.
\end{equation}

We optimize the learning direction of AFNM via calculating gradients of $\mathcal{L}_{trn}$, which is formulated as:
\begin{equation}
\begin{split}
    \theta_{F}' = \theta_{F} - \beta_1 \nabla_{\theta_{F}}
    \mathcal{L}_{trn}(\theta_{base},\theta_{F})
\end{split}
\end{equation}
\noindent \textbf{Meta-Test.}
Updated by meta-training, we utilize the batches from the remaining meta-test domains $D_{val}$ to simulate the real domain shifts.
AFNM optimized on $D_{trn}$ is also required to perform well on $D_{val}$ via updating it through below fomulation:
\begin{equation}
\begin{split}
    \mathcal{L}_{val}(\theta_{base},\theta_{F}')= \sum_{D_{val}} \mathcal{L}_{Cls} + \mathcal{L}_{Dep}+  \lambda_1 \mathcal{L}_{IDC} +  \lambda_2 \mathcal{L}_{ICS},
\end{split}
\end{equation}
\noindent \textbf{Meta-Optimization.}
In each iteration of meta-learning, we obtain $\mathcal{L}_{trn}$ and $\mathcal{L}_{val}$ from meta-train and meta-test for optimization, which is formulated as below:
\begin{equation}
    \theta_{F} \leftarrow \theta_{F} - \beta_2 \nabla_{\theta_{F}} (\mathcal{L}_{trn}(\theta_{base},\theta_{F}) + \mathcal{L}_{val}(\theta_{base},\theta_{F}')).
\end{equation}
In the above training strategy, $\theta_{F}$ is updated via meta-learning and $\theta_{base}$ is optimized in the normal training process, which not only improves the generalization of our method, but also facilitates the stability and efficiency of meta-learning.

\begin{algorithm}[t!]
    \caption{The optimization strategy of ANRL}
    \label{optimization_strategy}
    \KwData{$N$ source domains $D= [D_1,D_2,...,D_N]$}
    Initial parameters $\theta_{F}$ of AFNM and parameters $\theta_{base}$ of other modules.
    Determine learning rates $\beta_{1}$, $\beta_{2}$\ and hyper-parameters $\lambda_1$, $\lambda_2$;
    
    Shuffle all samples from different domains\;
    \For{$t$ in (1 : $N_{epoch})$} {
        
        \textbf{Normal train:} Sampling batch in $D$ \;
        $\mathcal{L}_{base}(\theta_{base},\theta_{F})= \sum_{D} \mathcal{L}_{Cls} + \mathcal{L}_{Dep}$ \;
        $\theta_{base} \leftarrow \theta_{base} - \beta_1 \nabla_{\theta_{base}}
        \mathcal{L}_{base}(\theta_{base},\theta_{F})$ \;
        
        \textbf{Meta-train:} Sampling batch in meta-train domains $D_{trn}$ \;
        
        $\mathcal{L}_{trn}(\theta_{base},\theta_{F}) = \mathcal{L}_{Cls(D_{trn})} + $ 
        $\mathcal{L}_{Dep(D_{trn})} + 
        \lambda_1 \mathcal{L}_{IDC(D_{trn})}+ 
        \lambda_2 \mathcal{L}_{ICS(D_{trn})}$ \;
        
        $\theta_{F}'=\theta_{F}-\beta_1\nabla_{\theta_{F}}\mathcal{L}_{trn}(\theta_{base},\theta_{F})$\;
        \textbf{Meta-test:} 
        Sampling batch in meta-test domains $D_{val}$ \;

        $\mathcal{L}_{val}(\theta_{base},\theta_{F}') = \mathcal{L}_{Cls(D_{val})}+$ 
        $\mathcal{L}_{Dep(D_{val})}+ 
        \lambda_1 \mathcal{L}_{IDC(D_{val})}+ 
        \lambda_2 \mathcal{L}_{ICS(D_{val})}$ \;
        
        \textbf{Meta-optimization:}
        
        $\theta_{F} \leftarrow \theta_{F} - \beta_2\nabla_{\theta_{F}}(\mathcal{L}_{trn}(\theta_{base},\theta_{F})+
        \mathcal{L}_{val}(\theta_{base},\theta_{F}'))$ \;
        
    }
    \Return Model parameters $\theta_{F}$ and $\theta_{base}$\;
\end{algorithm}

\begin{figure*}[htb]
	\centering
	\scalebox{1.0}{
	\includegraphics[width=1.0\linewidth]{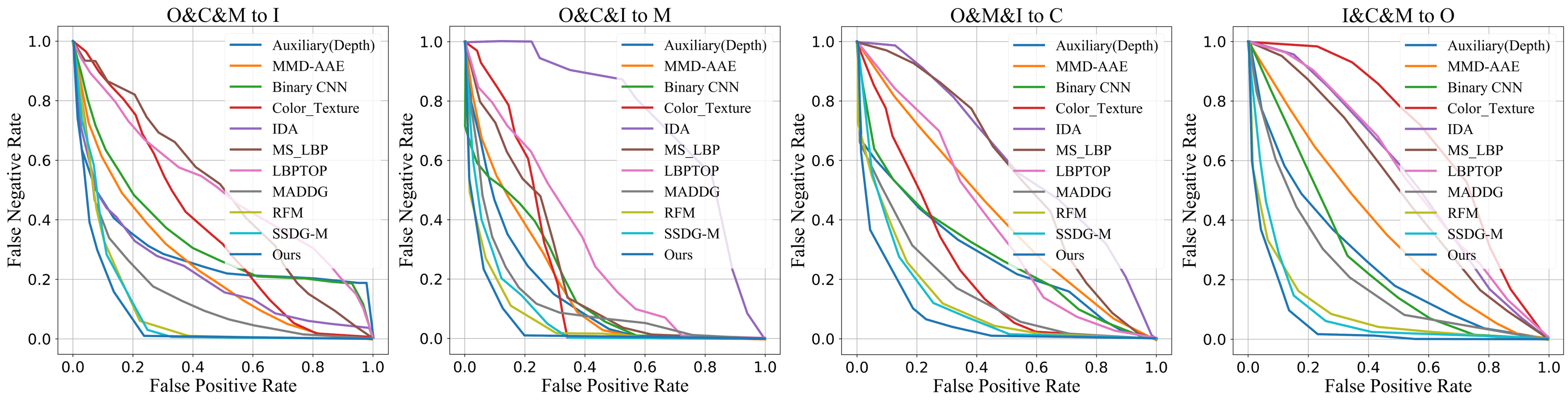}}
	\caption{ROC curves of four testing tasks for generalizable face anti-spoofing.}
	\label{fig:ROC}
\end{figure*}

\begin{table*}[t!]
    \caption{Comparison to other methods on four testing tasks for generalizable face anti-spoofing.}
    \begin{center}
	\setlength{\tabcolsep}{3mm}{
    \begin{tabular}{ccccccccc}
  \toprule
  \multirow{2}{*}{\textbf{Method}}&
    \multicolumn{2}{c}{\textbf{O\&C\&M to I}}&\multicolumn{2}{c}{\textbf{O\&C\&I to M}}&\multicolumn{2}{c}{\textbf{O\&M\&I to C}}&\multicolumn{2}{c}{\textbf{I\&C\&M to O}}\cr
    \cmidrule(lr){2-3} \cmidrule(lr){4-5} \cmidrule(lr){6-7} \cmidrule(lr){8-9}
    &HTER(\%)&AUC(\%)&HTER(\%)&AUC(\%)&HTER(\%)&AUC(\%)&HTER(\%)&AUC(\%)\cr
    \midrule
    MS\_LBP \cite{LBP03} &50.30&51.64&29.76&78.50&54.28&44.98&50.29&49.31\cr
    Binary CNN \cite{DeepBinary03} &34.47&65.88&29.25&82.87&34.88&71.94&29.61&77.54\cr
    IDA \cite{wen2015face} &28.35&78.25&66.67&27.86&55.17&39.05&54.20&44.59 \cr
    Color Texture \cite{other01} &40.40&62.78&28.09&78.47&30.58&76.89&63.59&32.71 \cr
    LBPTOP \cite{de2014face} &49.45&49.54&36.90&70.80&42.60&61.05&53.15&44.09 \cr
    Auxiliary(Depth Only)&29.14&71.69&22.72&85.88&33.52&73.15&30.17&77.61 \cr
    Auxiliary(All) \cite{additionInfo01} &27.6&-&-&-&28.4&-&-&-\cr
    MMD-AAE \cite{li2018domain} &31.58&75.18&27.08&83.19&44.59&58.29&40.98&63.08 \cr
    MADDG \cite{DG00} &22.19&84.99&17.69&88.06&24.5&84.51&27.98&80.02 \cr
    SSDG-M \cite{DG02} &18.21&\textbf{94.61}&16.67&90.47&23.11&85.45&25.17&81.83\cr
    RFM \cite{DG01} &17.3&90.48&13.89&93.98&20.27&88.16&16.45&91.16 \cr
    \midrule
    \textbf{Ours}&\textbf{16.03}&91.04&\textbf{10.83}&\textbf{96.75}&\textbf{17.85}&\textbf{89.26}&\textbf{15.67}&\textbf{91.90}\cr
    \bottomrule
`    \end{tabular}}
    \end{center}
    \vspace{-4mm}
    \label{tab:anti_1}
\end{table*}

\begin{table}[t!]
\caption{Comparison to face anti-spoofing methods with limited source domains.}
\begin{center}
\scalebox{0.95}{
\setlength{\tabcolsep}{2mm}{
\begin{tabular}{ccccc}
  \toprule
  \multirow{2}{*}{\textbf{Method}}&
    \multicolumn{2}{c}{\textbf{M\&I to C}}&\multicolumn{2}{c}{\textbf{M\&I to O}}\cr
    \cmidrule(lr){2-3} \cmidrule(lr){4-5}
    &HTER(\%)&AUC(\%)&HTER(\%)&AUC(\%)\cr
    \midrule
    MS\_LBP~\cite{LBP03} &51.16&52.09&43.63&58.07\cr
    IDA~\cite{wen2015face} &45.16&58.80&54.52&42.17\cr
    Color Texture~\cite{other01} &55.17&46.89&53.31&45.16 \cr
    LBPTOP~\cite{de2014face} &45.27&54.88&47.26&50.21 \cr
    MADDG~\cite{DG00} &41.02&64.33&39.35&65.10 \cr
    SSDG-M~\cite{DG02} &31.89&71.29&36.01&66.88 \cr
    \midrule
    \textbf{Ours}&\textbf{31.06}&\textbf{72.12}&\textbf{30.73}&\textbf{74.1} \cr
    \bottomrule
    \end{tabular}}}
    \end{center}
    \vspace{-4mm}
    \label{tab:anti_2}
\end{table}

\section{Experiments}
\subsection{Experimental Settings}

\noindent \textbf{Datasets.} 
Following the setting of MADDG \cite{DG00}, we evaluate the effectiveness of our method with four public face anti-spoofing datasets, OULU-NPU~\cite{Oulu} (denoted as O), CASIA-FASD~\cite{CASIA} (denoted as C), Idiap Replay-Attack~\cite{Replay} (denoted as I), and MSU-MFSD~\cite{wen2015face} (denoted as M).
Concretely, We randomly select three datasets from them as source domains and the left one is treated as target domain, which is unavailable during the training process. 
Thus, we have four testing tasks in total: O\&C\&I to M, O\&M\&I to C, O\&C\&M to I, and I\&C\&M to O. For significant domain shifts (\textit{e.g.}, background, illustration, material and \textit{etc}.) exist under the cross-dataset scenarios, the domain generalization for face anti-spoofing is a challenging task.

\noindent \textbf{Implementation Details.} Our method is implemented via PyTorch~\cite{pytorch} on $11G$ NVIDIA $2080$Ti GPUs with Linux OS and trained with Adam optimizer~\cite{Kingma2015AdamAM}. We utilize the RGB and HSV channels of each image, which means the input size of our method is $256\times256\times6$ to extract features following the network architecture in RFM~\cite{DG01}. For training, the hyper-parameters $\lambda_{1}$ and $\lambda_{2}$ are set to $0.1$ and $0.01$ respectively. Both learning rates $\beta_{1}$ and $\beta_{2}$ are set to $0.001$. The momentum factor $\gamma$ is set to $0.9$. We strictly follow the popular evaluation metrics, which contain Half Total Error Rate (HTER) and the Area Under Curve (AUC).

\subsection{Experimental Results}
As shown in Table~\ref{tab:anti_1} and Figure~\ref{fig:ROC}, we make the following observations. (1) DG-based face anti-spoofing methods~\cite{DG00, DG01} perform better than conventional methods~\cite{wen2015face, other01}. This proves that the distribution of the target domain is different from source domains, while the conventional methods focus on the differentiation cues that only fit source domains. 
(2) Our method outperforms these DG-based methods under four test settings, which demonstrates the effectiveness of ANRL. This is because all these methods endeavor to construct a compact and generalized feature space only based on BN, which is intrinsically fragile for the domain shift, resulting in the performance degradation. However, our ANRL leverages IN to filter out the domain bias, and then combines both features with balance factors to take advantage of them. Moreover, due to the variations between samples, the balance factor is customized based on the information of each sample for the most suitable estimation. Last but not least, to further regularize the representation space, we utilize DCC to reduce the disturbance of domain bias and enlarge the distance between real and fake samples.

\noindent \textbf{Limited Source Domains.} As illustrated in Table~\ref{tab:anti_2}, we also evaluate our method with extremely limited source domains (\textit{i.e.}, only two source datasets). Specifically, MSU and Idiap databases are selected as the source domains for training and the remaining two, \textit{i.e.}, CASIA and OULU respectively, are used as the target domains for testing. Our proposed method achieves the best performance in this more challenging case, which powerfully verifies its generalizability on unseen target domains. 

\begin{figure*}[t!]
	\centering
    \includegraphics[width=1.0\linewidth]{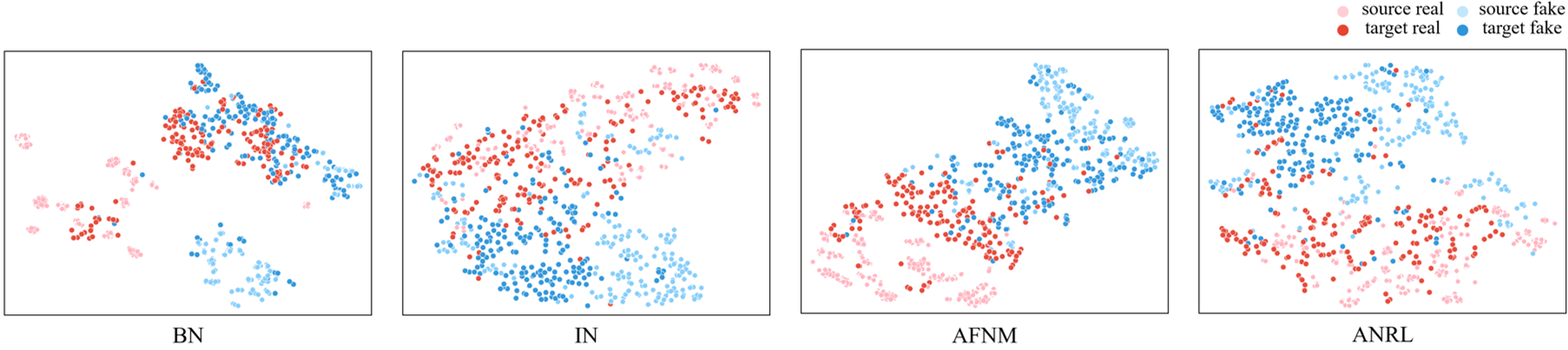}
	\caption{The t-SNE visualization of features extracted by different variants on task I\&C\&M to O with meta-learning. Since BN may be corrupted by domain gap and IN may loss some discriminative information, our ANRL contributes to construct a domain-agnostic and spoofing-discriminative representation space for the best performance.}
	\label{fig:tSNE}
\end{figure*}

\subsection{Ablation Study}
\noindent \textbf{Effectiveness of Different Components.} As shown in Table~\ref{tab:anti_4}, we validate the effectiveness of each module on task I\&C\&M to O and find out the following observations. (1) Since meta-learning can facilitate the optimization of model parameters to obtain more generalized representation, it indeed improves the performance compared to AFNM without meta-learning in the first row. (2) Besides IN can filter out some domain bias, $L_{IDC}$ is furthermore devised to align samples from different domains to eliminate the negative impacts of domain bias, achieving better performance. (3) To improve the discriminative ability, we design $L_{ICS}$ to push samples of different categories away, leading to better results. (4) ANRL which combines all above complementary modules for a domain-agnostic and spoofing-discriminative representation space, yields the best performance compared to the other variants. 

\begin{table}[t]
    \caption{Evaluation of different components in our method on the task \textbf{I\&C\&M to O}.}
    \begin{center}
    \setlength{\tabcolsep}{4mm}{
    \begin{tabular}{ccccc}
      \toprule
      \multirow{2}{*}{\textbf{Meta}}&\multirow{2}{*}{\bm{$L_{ICS}$}}&\multirow{2}{*}{\bm{$L_{IDC}$}}&
        \multicolumn{2}{c}{\textbf{I\&C\&M to O}}\cr
        \cmidrule(lr){4-5}
        &&&HTER(\%)&AUC(\%)\cr
        \midrule
        & & &19.23 &87.98\cr
        \checkmark& & &17.61 &89.30\cr
        \checkmark& &\checkmark &16.75 &90.87\cr
        \checkmark &\checkmark & &16.43 &91.23\cr
        \textbf{\checkmark} &\textbf{\checkmark} &\textbf{\checkmark} &\textbf{15.67} &\textbf{91.90} \cr
        \bottomrule
        \end{tabular}}
        \end{center}
        \label{tab:anti_4}
\end{table}

\begin{table}[t]
    \caption{Comparison to the other normalization methods trained with meta learning for face anti-spoofing domain generalization on the task \textbf{I\&C\&M to O}.}
    \begin{center}		
    \setlength{\tabcolsep}{8mm}{
    \begin{tabular}{ccc}
   \toprule
   \multirow{2}{*}{\textbf{Method}}&
    \multicolumn{2}{c}{\textbf{I\&C\&M to O}}\cr
    \cmidrule(lr){2-3}
    &HTER(\%)&AUC(\%)\cr
    \midrule
    BN &23.37 &82.42 \cr
    IN &21.86 &85.76 \cr
    IN-BN-half &20.65 &86.23 \cr
    BIN~\cite{BIN01} &19.72 &87.39 \cr
    IBN~\cite{Pan2018TwoAO} &20.05 &86.88 \cr
    \textbf{AFNM} &\textbf{17.61} &\textbf{89.30} \cr

    \bottomrule
    \end{tabular}}
    \end{center}
    \label{tab:anti_3}
\end{table}

\begin{figure}[t!]
	\centering
	\includegraphics[width=1.0\linewidth]{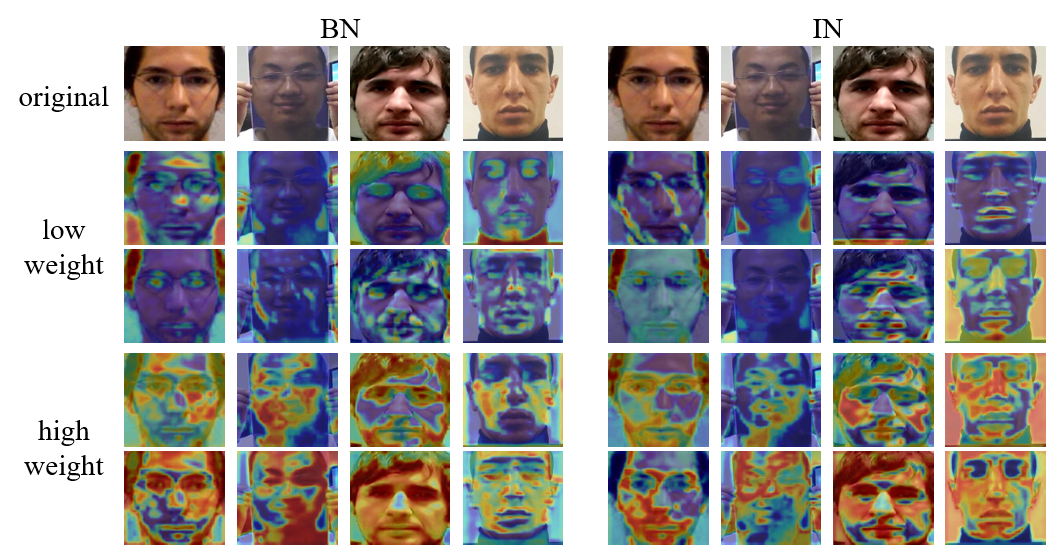}
	\caption{Feature maps of low-weight and high-weight channels of IN and BN from AFNM3 on task I\&C\&M to O.}
	\label{fig:feature_map_channel}
\end{figure}

\begin{figure}[t!]
	\centering
	\includegraphics[width=1.0\linewidth]{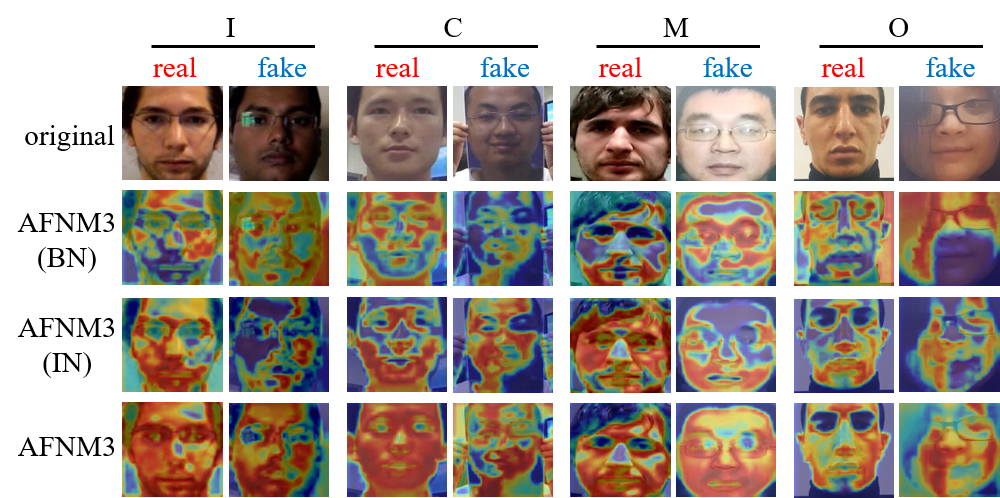}
	\caption{Weighted feature maps of BN, IN along channels and their fusion from AFNM3 on task I\&C\&M to O.}
	\label{fig:feature_map}
\end{figure}

\noindent \textbf{Effectiveness of AFNM.} To further verify the effectiveness of AFNM, as shown in Table~\ref{tab:anti_3} and Figure~\ref{fig:tSNE}, we compare it with other normalization methods trained with meta-learning on task I\&C\&M to O and make the following observations. (1) Due to BN is sensitive to domain drift, it does not perform very well on the target domain, though the good performance on source domains. (2) Since IN filters out some domain variations, it attains better results on the target domain, although inferior to BN on source domains. (3) IN-BN-half simply combines features from IN and BN with the fixed balance factor 0.5 for all channels evenly, which achieves better performance. (4) BIN dynamically learns the shared balance factor at channel level, superior to IN-BN-half. IBN concatenates the features from BN and IN layers along the channels, acquiring comparable performance. (5) Due to the different degrees of domain information in samples, AFNM adaptively fuses features from BN and IN with the costumed balance factor for each sample, getting better results. (6) By adding DCC to guide the learning of AFNM, ANRL further reduces differences between domains and increases margin between classes, achieving the best performance.

\section{Analysis}
\subsection{Feature Map Analysis}
To provide more insights on how ANRL commits to learning a generalizable representation combined with the features from IN and BN, we visualize the feature maps from AFNM3 in Figure~\ref{fig:framework} on the task I\&C\&M to O. Firstly, as shown in Figure~\ref{fig:feature_map_channel}, it is obvious that selected high-weight channels of IN or BN all focus more on face regions for intrinsic spoofing cues, which can generalize better on unseen target domains. Meanwhile, the low-weight channels pay more attention to specific cues related to source domains, \textit{e.g.}, hands and background, which are not transferable. Moreover, as shown in Figure~\ref{fig:feature_map}, after selecting the discriminative channels of IN and BN, the weighted feature maps AFNM3(BN) and AFNM3(IN) are complementary. Therefore, AFNM3 adaptively combines both of them to focus more completely on facial area for better generalization.

\begin{figure}[t!]
	\centering
	\includegraphics[width=1.0\linewidth]{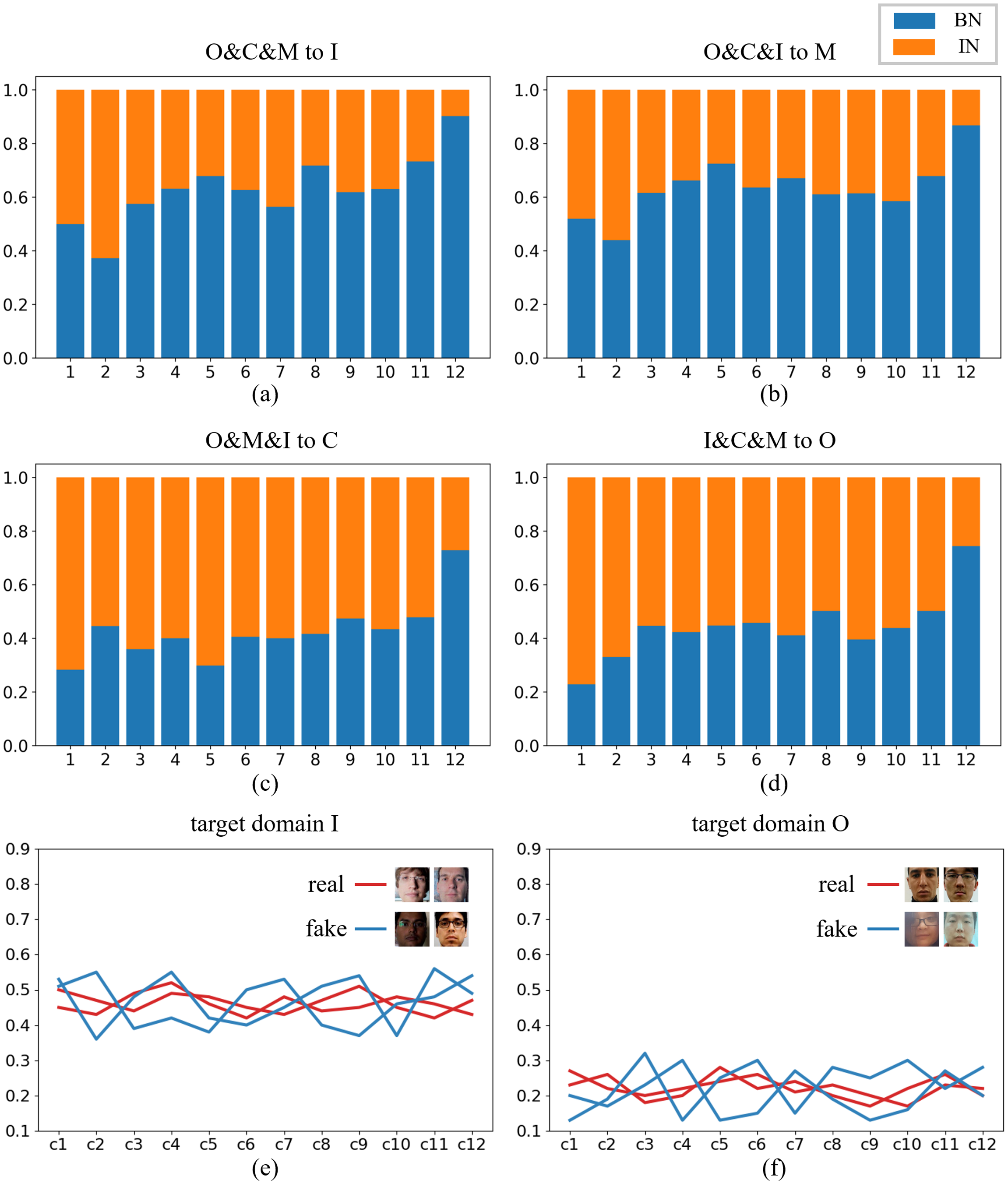}
	\caption{(a)-(d) The mean of balance factor $\bm{\alpha}$ of each layer for the feature combination on four testing tasks. The numbers represent the corresponding layers. (e)-(f) Some channels of $\bm{\alpha}$ of $\bm{layer1}$ for different samples on target domain I and O. The different channels are denoted by $\bm{c1}$-$\bm{c12}$.}
	\vspace{-3mm}
	\label{fig:balance_factor}
\end{figure}

\subsection{Balance Factor Analysis}

\noindent \textbf{Mean of Balance Factor.} To understand how balance factors influence the normalized representation, we investigate the mean of balance factor $\alpha$ of different layers in the feature extractor. It is noted that all balance factors are initialized to 0.5. As shown in Figure~\ref{fig:balance_factor} (a)-(d), for the low-level features contain more style variations associated with domain, $\alpha$ in the shallow layers tends to zero, indicating that the low layers prefer IN to mitigation the domain discrepancy across different datasets. While since the high-level features are prone to be utilized for classification, which coincides with the role of BN, $\alpha$ in the high layers tends to one. Moreover, we find out that the degree of IN utilized on different tasks varies from each other. The task O\&M\&I to C and I\&C\&M to O leverage IN more compared to O\&C\&M to I and O\&C\&I to M, this is probably due to the domain variance in multiple source domains is too large on the previous two tasks and therefore IN is urgently needed to filter out the variance for generalizable FAS.

\begin{figure}[t!]
	\includegraphics[width=0.9\linewidth]{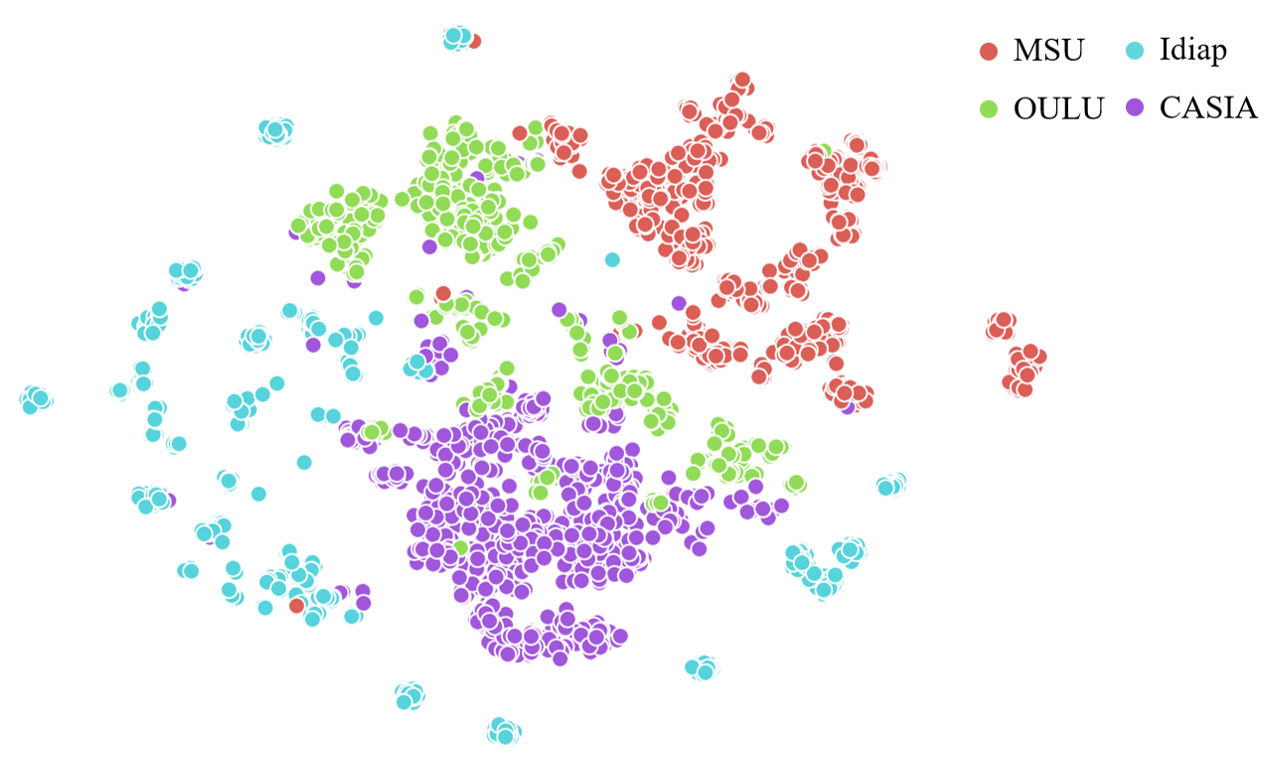}
	\caption{The t-SNE visualization of features of four datasets extracted by ResNet pre-trained on ImageNet.}
	\label{fig:mean_std}
\end{figure}

\noindent \textbf{Adaptability of Balance Factor.} Since the domain information of each sample is different, the best balance factors $\alpha$ of them are not the same. As shown in Figure~\ref{fig:balance_factor} (e)-(f), we select some channels of $\alpha$ with big variances in $layer1$ of feature extractor, and we observe that the balance values $\alpha_{c1}\sim\alpha_{c12}$ of different samples are significantly various, which proves that ANRL can adaptively determine the most suitable $\alpha$ for each sample. Furthermore, because fake samples are more diverse than real ones, their variance is larger.

\noindent \textbf{Further Exploration.} To further explore why ANRL leverages more IN on some specific tasks, as shown in Figure \ref{fig:mean_std}, we utilize t-SNE~\cite{Maaten2008VisualizingDU} to visualize the feature distribution of four datasets. we leverage ResNet~\cite{He2016DeepRL} pre-trained on ImageNet to extract relatively objective features, and find out that the difference between OULU and CASIA is relatively smaller compared to the one between Idiap and MSU. Therefore, we speculate that when the task simultaneously contains Idiap and MSU as source domains, due to the large discrepancy of them, ANRL will automatically learn to utilize more IN to filter out domain biases.

\section{Conclusion}
In this paper, we propose the Adaptive Normalized Representation Learning (ANRL) framework to obtain a domain-agnostic and discriminative representation via adaptively selecting features from BN and IN.
Concretely, we devise Adaptive Feature Normalization Module (AFNM) to estimate the customized combination factor for each sample, which is aware of the distinction among samples. 
Furthermore, to guide the learning of factors, Dual Calibration Constraints are proposed, including contains Inter-Domain Compatible Loss and Inter-Class Separable Loss.
They cooperatively provide a better optimization direction to update AFNM via meta-learning from the perspective of domain and class, leading to more generalizable representation.
Extensive experiments on public datasets demonstrate the effectiveness of our proposed method.

\section{Acknowledgments}
  This work was supported by National Natural Science Foundation of China (No. 61972157), National Key Research and Development Program of China (No. 2019YFC1521104), Shanghai Municipal Science and Technology Major Project (2021SHZDZX0102), Zhejiang Lab (No. 2020NB0AB01).
\bibliographystyle{ACM-Reference-Format}
\balance
\bibliography{paper_ref}

\end{document}